\documentclass[letterpaper, 10 pt, conference]{ieeeconf}  
\IEEEoverridecommandlockouts                              
\usepackage{graphics} 
\usepackage{epsfig} 
\usepackage[OT1]{fontenc} 
\usepackage{amsmath} 
\usepackage{amssymb}  
\usepackage{color}
\usepackage{algorithm}
\usepackage[noend]{algpseudocode}
\usepackage{hyperref}
\usepackage{pdfpages}

\usepackage{caption}
\DeclareCaptionFormat{twolinetab}{#1#2#3}
\usepackage{url}

\usepackage{amsthm}
\newtheorem{thm}{Theorem}
\newtheorem{lem}{Lemma}
\newtheorem{prop}{Proposition}
\newtheorem{cor}{Corollary}
\theoremstyle{definition}
\newtheorem{ass}{Assumption}
\newtheorem{defn}{Definition}
\newtheorem{exmp}{Example}
\newtheorem{rem}{Remark}

\newcommand{\wip}{false}
\newcommand{\proofs}{true}

\newcommand{\IfCustom}[2][true]{
\ifthenelse{\equal{#1}{true}}
{#2}
{}
}

\title{\LARGE \bf
Data-driven abstractions via adaptive refinements and a Kantorovich metric 
}

\author{Adrien Banse, Licio Romao, Alessandro Abate and Rapha\"{e}l M. Jungers
\thanks{R. M. Jungers is a FNRS honorary Research Associate. This project has received funding from the European Research Council (ERC) under the European Union’s Horizon 2020 research and innovation program under grant agreement No 864017 - L2C. R. M. Jungers is also supported by the Walloon Region and the Innoviris Foundation. He is currently on sabbatical leave at Oxford University, Department of Computer Science, Oxford, UK.
Adrien Banse is supported by the French Community of Belgium in the framework of a FNRS/FRIA grant.
Adrien Banse and Rapha\"{e}l M. Jungers are with ICTEAM, UCLouvain. E-mail adresses: \texttt{\{adrien.banse, raphael.jungers\}@uclouvain.be}.
Licio Romao and Alessandro Abate are with the Department of Computer Science, Oxford University. E-mail adresses: \texttt{\{licio.romao, alessandro.abate\}@cs.ox.ac.uk}.
}
}

\begin{document}

\maketitle

\thispagestyle{empty}
\pagestyle{empty}
\IfCustom[\wip]{
   \thispagestyle{plain}
   \pagestyle{plain}
}

\begin{abstract}
   We introduce an adaptive refinement procedure for smart and scalable abstraction of dynamical systems. Our technique relies on partitioning the state space depending on the observation of future outputs. However, this knowledge is dynamically constructed in an adaptive, asymmetric way. In order to learn the optimal structure, we define a Kantorovich-inspired metric between Markov chains, and we use it to guide the state partition refinement. Our technique is prone to data-driven frameworks, but not restricted to.

   We also study properties of the above mentioned metric between Markov chains, which we believe could be of broader interest. We propose an algorithm to approximate it, and we show that our method yields a much better computational complexity than using classical linear programming techniques.
\end{abstract}


\section{Introduction}


Feedback control of dynamical systems is at the core of several techniques that have caused tremendous impact in several industries, being essential to important advancements in e.g. aerospace and robotics. Traditionally, these control techniques were model-based, relying on a complete mathematical model to perform controller design. With recent technological advancements, however, where a vast amount of data can be collected online or offline, the interest within the control community to study methods that leverage available data for feedback controller design has been reignited \cite{Persis2020,Wang2021,Berberich2021,Banse2022}.

In this paper, we focus on data-driven techniques for building \emph{abstractions of dynamical systems.} We call these \emph{data-driven abstractions}. Abstraction methods create a symbolic model \cite{Lind1995,Schaft2004} that approximates the behavior of the original (the ``concrete'') dynamics in a way that controllers designed for such a symbolic representation can be refined to a valid controller for the original dynamics \cite{Tabuada2009}. 
Several recent research efforts started exploring the possibility of generating abstractions for dynamical systems from observations of the latter \cite{pmlr-v211-banse23a,Devonport2021,Coppola2023,Lavaei2023, Badings2023}. 

In \cite{pmlr-v211-banse23a}, we show that memory-based Markov models can be built from trajectory data. Memory has been classically used as a tool to mitigate non-Markovian behaviors of the original dynamics \cite{Majumdar2020,Schmuck2014}, a feature also explored in recent papers \cite{Majumdar2020,Coppola2023}. Increasing memory allows us to create more precise representations of the original dynamics using Markov decision processes (MDPs) or Markov chains. 
Despite promising results,  \cite{pmlr-v211-banse23a} does not offer an adaptive mechanism to compute the generated abstraction, and thus it faces the curse of dimensionality, as the number of possible observations grows exponentially with the memory length. In this paper, we further develop the techniques in \cite{pmlr-v211-banse23a} by proposing an adaptive state space partitioning to mitigate the curse of dimensionality.

\begin{figure}[ht!]
    \centering\includegraphics[width = 0.5\columnwidth]{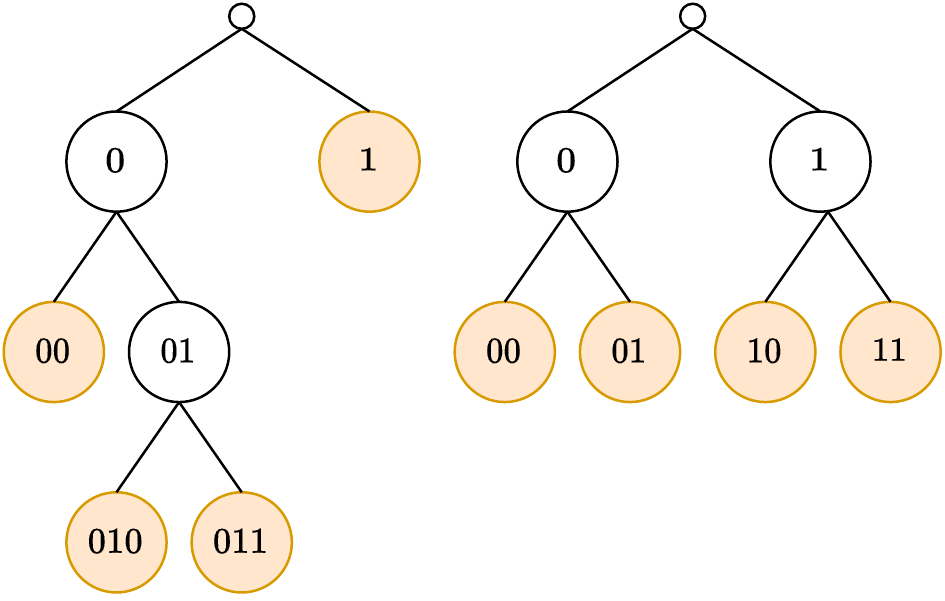}
    \caption{Difference between an adaptive and a brute-force approach. On the left, an adaptive approach: only some of the possible observations are expanded. On the right, a brute-force approach: every observation is expanded. The two resulting abstractions have four states, but the adaptive one provides a better abstraction (see Example~\ref{exmp:memory}).}
    \label{fig:adaptive}
\end{figure}

A key contribution in this paper is the construction of a novel metric between two Markov chains; this metric is then exploited to adaptively increase memory in certain regions of the state space, in view of taming the complexity of the generated abstraction. An illustration of the difference between these two approaches is depicted in Figure~\ref{fig:adaptive}. As opposed to \cite{pmlr-v211-banse23a}, where states of the chain are associated with past memory, the abstractions we construct in this paper are based on forward memory. In order to define a metric  between two Markov models, we leverage the \emph{Kantorovich} metric (also known as the Wasserstein or Earth's mover distance) between the induced probability on words of a fixed length and let the word length go to infinity. To define the Kantorovich metric, we equip the space of words with the \emph{Cantor distance}, which is classically used in symbolic dynamics \cite{Lind1995, Fogg2002}. We argue by means of numerical experiments that such a Kantorovich metric is natural and meaningful for control problems. We also show that the proposed metric is a well-defined and intuitive notion of similarity between Markov chains, and propose an algorithm for its computation that has better computational complexity over a naive application of linear programming techniques.

We believe that the proposed Kantorovich metric could be of much broader interest. Indeed, computing metrics between Markov models has been an active research topic within the computer science community \cite{Deng2009}. Our construction on the metric between Markov chains resembles the one presented in \cite{Rached2004}, however with another distance. In \cite{Kiefer2018} computability and complexity results are shown for the total variation metric. Kantorovich metrics for Markov models have been studied in \cite{Desharnais2004,Breugel2007,Madras2010,Rudolf2018}, but their underlying distance is different from ours. Our choice of the Cantor distance is crucial both for computational aspects and for building smart abstractions of dynamical systems.

Summarizing, our main contribution is threefold. First, we propose a new metric to measure distance between Markov models. Second, we develop an efficient algorithm that approximates arbitrarily well the proposed metric. Third, we exploit the proposed metric to adaptively improve abstractions in specific regions of the state space.









\paragraph*{Outline} The rest of this paper is organized as follows. In Section~\ref{sec:kantorovich}, we introduce the Kantorovich metric between two Markov chains, and propose an efficient algorithm to approximate it with arbitrary precision. In Section~\ref{sec:abstractions}, we apply this metric to build abstractions of dynamical systems using a greedy strategy that leads to the refinement of the state-space partitioning. We also demonstrate the quality of our procedure on an example. 

\paragraph*{Notations} Let $\mathcal{A}$ be a finite alphabet. We denote the set of $n$-long sequences of this alphabet by $\mathcal{A}^n$, and the set of countably infinite sequences by $\mathcal{A}^*$. The symbol $\Lambda$ stands for the empty sequence and, for any $w_1 \in \mathcal{A}^{n_1}$, $w_2 \in \mathcal{A}^{n_2}$, the sequence $w_1w_2 \in \mathcal{A}^{n_1 + n_2}$ is the concatenation of $w_1$ and $w_2$. Let $c(x)$ be a number of operations with respect to some attributes $x$. We say that an algorithm has a computational complexity $\mathcal{O}(f(x))$ if there exists $M > 0$, $x_0$ such that, for all $x \geq x_0$, $c(x) \leq Mf(x)$. For any bounded set $X \subset \mathbb{R}^d$, let $\sigma(X)$ be the induced $\sigma$-algebra of $X$, and $\lambda$ be the Lebesgue measure on $\mathbb{R}^d$, then $\lambda_X : \sigma(X) \to [0, 1]$ is defined as $\lambda_X(A) = \lambda(A) / \lambda(X)$. Finally, for any set $X$ and function $F$, the set $F(X) = \{F(x) : x \in X\}$.

\section{A Kantorovich metric between Markov chains} \label{sec:kantorovich}

In this section, a new notion of metric between Markov chains is defined. In Section~\ref{sec:abstractions} below, 
Markov chains will be used to represent abstractions of dynamical systems, and this distance will be used as a tool to construct adaptive abstractions. The present section, however, is concerned with Markov chains in their full generality.

\subsection{Preliminaries}
Using a similar formalism as in \cite{pmlr-v211-banse23a}, we define a labeled Markov chain as follows.
\begin{defn}[Markov chain] \label{def:markov}
   A \emph{Markov chain} is a 5-tuple $\Sigma = (\mathcal{S}, \mathcal{A}, P, \mu, L)$, where $\mathcal{S}$ is a finite set of \emph{states}, $\mathcal{A}$ is a finite \emph{alphabet}, $P$ is the \emph{transition matrix} on $\mathcal{S} \times \mathcal{S}$, $\mu$ is the \emph{initial measure} on $\mathcal{S}$, and $L: \mathcal{S} \to \mathcal{A}$ is a \emph{labelling function}.
\end{defn}
In Definition~\ref{def:markov}, the entry of the transition matrix $P_{s, s'}$ represents the probability $\mathbb{P}(X_{k + 1} = s' | X_{k} = s)$. The labelling $L$ induces a partition of the states. Consider the equivalence relation on $\mathcal{S}$ defined as $s \sim s'$ if and only if $L(s) = L(s')$. For any $a \in \mathcal{A}$, the notion of \emph{equivalent classes} is defined as 
\begin{equation} \label{eq:equiv_markov}
   [a] = \{s \in \mathcal{S}: L(s) = a\}.
\end{equation} 
We also define the \emph{behavior of a Markov chain} $\mathcal{B}(\Sigma) \subseteq \mathcal{A}^*$ as follows. A sequence $w^* = (a_1, a_2, \dots) \in \mathcal{B}(\Sigma_\mathcal{W})$ if there exists $s_1, s_2, \dots \in \mathcal{S}$ such that $\mu_{s_1} > 0$, $P_{s_i, s_{i+1}} > 0$ and $L(s_i) = a_i$. 

In the present work, we focus on a notion of metric between probabilities on label sequences. Let $w = (a_1, \dots, a_n)$ be a $n$-long sequence of labels, and define $p^n : \mathcal{A}^n \to [0, 1]$ as
\begin{equation} \label{eq:distribution}
   p^n(w)
   = 
   \sum_{s_1 \in [a_1]} \mu_{s_1} 
   \sum_{s_2 \in [a_2]} P_{s_1, s_2} 
   \dots
   \sum_{s_n \in [a_n]} P_{s_{n-1}, s_n}, 
\end{equation}
that is the probability induced by the Markov chain on $n$-long sequences.

\begin{rem} \label{rem:prob_complexity} 
Classical procedures are well-known in the literature allowing to compute the probabilities $p^n$ for increasing $n$, with a complexity proportional to $|\mathcal{S}|^2$ at every step \cite{Rabiner1989}.
\end{rem}


We endow the set of $n$-long sequences of labels with the Cantor distance $d_B$.
\begin{defn}[Cantor's distance, \cite{Fogg2002}] \label{def:baire}
   The \emph{Cantor distance} $d_B: \mathcal{A}^n \times \mathcal{A}^n \to \mathbb{R}$ is defined as 
      $d_B(w_1, w_2) = 2^{-l}$, 
   where $l$ is the length of the longest common prefix. In other words, let  $w_1 = (a_1, \dots, a_n)$ and $w_2 = (b_1, \dots, b_n)$, then $d_B(w_1, w_2) = 2^{-l}$, where $l = \inf\{ k : a_k \neq b_k \}$.
\end{defn}
\IfCustom[\proofs]{
   \begin{rem}\label{rem:ultra}
      It is well-known that the Cantor distance is an ultrametric. It means that is satisfies the strong triangular inequality 
   \begin{equation}
      d_B(w_1, w_3) \leq \max\{d_B(w_1, w_2), d_B(w_2, w_3)\}.
      \end{equation}
      This property will be crucial in our developments.
   \end{rem}
}


\subsection{The Kantorovich metric}

\begin{figure}[ht!]
   \centering
   \includegraphics[scale=0.55]{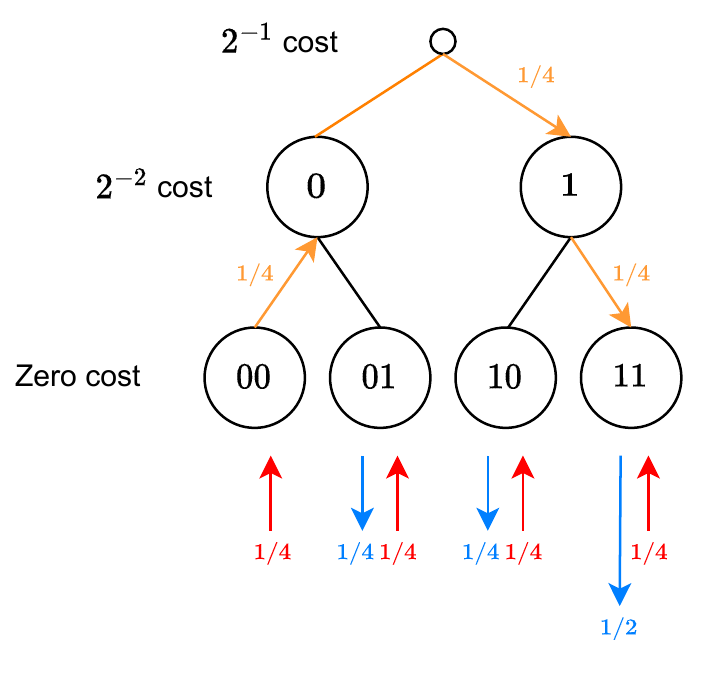}
   \caption{Interpretation of the Kantorovich distance as an optimal transport problem. In this example, the alphabet $\mathcal{A} = \{0, 1\}$, $p_1^2(w) = 1/4$ for all $w \in \mathcal{A}^2$, and $p_2^2(00) = 0$, $p_2^2(01) = p_2^2(10) = 1/4$, and $p_2^2(11) = 1/2$. One can see that the optimal way to satisfy the demands $p_2^n$ with the supplies $p_1^n$ is to move $1/4$ of probability mass from $00$ to $11$, that is $\pi^2(00, 11) = 1/4$. Since $d_B(00, 11) = 1/2$, the Kantorovich distance is $K(p^2_1, p^2_2) = 1/8$.}
   \label{fig:OT}
\end{figure}

Consider two Markov chains $\Sigma_1 = (\mathcal{S}_1, \mathcal{A}, P_1, \mu_1, L_1)$ and $\Sigma_2 = (\mathcal{S}_2, \mathcal{A}, P_2, \mu_2, L_2)$ defined on the same alphabet $\mathcal{A}$. For a fixed $n$, they generate the distributions $p^n_1$ and $p^n_2$ on the metric space $(\mathcal{A}^n, d_B)$ as described in \eqref{eq:distribution}.
\begin{defn}[Kantorovich metric]
   The \emph{Kantorovich metric} between the probability distributions $p_1^n$ and $p_2^n$ is given by 
   \begin{equation} \label{eq:kantorovich}
      K(p_1^n, p_2^n) = 
      \min_{\pi^n \in \Pi(p_1^n, p_2^n)}
      \sum_{w_1, w_2 \in \mathcal{A}^n} d_B(w_1, w_2)\pi^n(w_1, w_2), 
   \end{equation}
   where $\Pi(p_1^n, p_2^n)$ is the set of \emph{couplings} of $p_1^n$ and $p_2^n$, which contains the joint distributions $\pi^n: \mathcal{A}^n \times \mathcal{A}^n \to [0, 1]$ whose marginal distributions are $p_1^n$ and $p_2^n$, that is, 
   \begin{equation}\label{eq:linear_constraints}
    \begin{gathered} 
       \forall w_1, w_2 \in \mathcal{A}^n: \, \pi^n(w_1, w_2) \geq 0,\\
       \forall w_1 \in \mathcal{A}^n: \, \sum_{w_2 \in \mathcal{A}^n} \pi^n(w_1, w_2) = p_1^n(w_1), \\
       \forall w_2 \in \mathcal{A}^n: \, \sum_{w_1 \in \mathcal{A}^n} \pi^n(w_1, w_2) = p_2^n(w_2).
    \end{gathered}
    \end{equation}
\end{defn}
The Kantorovich metric is often interpreted as an optimal transport problem. Indeed one can see problem~\eqref{eq:kantorovich} as the problem of finding the optimal way to satisfy ``demands'' $p_2^{n}$ with ``supplies'' $p_1^n$, where the cost of moving $\pi^n(w_1, w_2)$ probability mass from $w_1$ to $w_2$ amounts to  $\pi^n(w_1, w_2) d_B(w_1, w_2)$. An illustration is provided in Figure~\ref{fig:OT}.

A na\"ive computation of $K(p_1^n, p_2^n)$ in \eqref{eq:kantorovich} entails solving a linear program. However, standard techniques, such as interior point methods and network simplex result in some cases in a complexity of $\mathcal{O}(n|\mathcal{A}|^{3n}\log(|\mathcal{A}|))$, and therefore scale very poorly with the number labels. In this section, we show that it is possible to compute $K(p_1^n, p_2^n)$ in $\mathcal{O}(|\mathcal{S}|^2|\mathcal{A}|^{n+1})$ operations. 

\IfCustom[\proofs]{
   We first present two lemmata that will be useful for our purpose.
   \begin{lem} \label{lem:min}
   For any $n \geq 1$, let $\pi^n$ be the solution of \eqref{eq:kantorovich}. For all $w \in \mathcal{A}^n$,
   \begin{equation}
      \pi^n(w, w) = \min\{p_1^n(w), p_2^n(w)\}.
   \end{equation}
   \end{lem}

   \begin{lem} \label{lem:raph}
      For any $n \geq 1$, let $\pi^{n+1}$ be the solution of \eqref{eq:kantorovich}. Then, for all $w \in \mathcal{A}^n$ such that $p^n_1(w) > p^n_2(w)$, then
      \begin{equation}
         \begin{aligned}
            \sum_{\substack{w' \in \mathcal{A}^n\\w' \neq w}} \sum_{a_1, a_2 \in \mathcal{A}} \pi^{n+1}(wa_1, w'a_2) &= p^n_1(w) - p^n_2(w),  \\
            \sum_{\substack{w' \in \mathcal{A}^n\\w' \neq w}} \sum_{a_1, a_2 \in \mathcal{A}} \pi^{n+1}(w'a_1, wa_2) &= 0, 
         \end{aligned}
      \end{equation} 
      and for all $w_1 \in \mathcal{A}^n$ such that $p_1^n(w_1) \leq p_2^n(w_1)$,
      \begin{equation}
         \begin{aligned}
            \sum_{\substack{w' \in \mathcal{A}^n\\w' \neq w}} \sum_{a_1, a_2 \in \mathcal{A}} \pi^{n+1}(wa_1, w'a_2) &= 0,  \\
            \sum_{\substack{w' \in \mathcal{A}^n\\w' \neq w}} \sum_{a_1, a_2 \in \mathcal{A}} \pi^{n+1}(w'a_1, wa_2) &= p^n_2(w) - p^n_1(w).
         \end{aligned}
      \end{equation} 
   \end{lem}
   }

We present in Theorem~\ref{thm:recursion} a key result for writing an efficient algorithm. Due to space constraints, all the proofs are in the appendices of this paper.

\begin{thm} \label{thm:recursion}
   For any $n \geq 1$, let $\pi^{n}$ be the solution of \eqref{eq:kantorovich}. Then the following holds:
   \begin{equation} \label{eq:iterative_prop}
   \begin{aligned}
      &K(p_1^{n+1}, p_2^{n+1}) = K(p_1^n, p_2^n) \\
      & + 2^{-(n+1)} \sum_{w \in \mathcal{A}^{n}}
      \left[
         r(w) 
         -
         \sum_{a \in \mathcal{A}}
         r(wa) 
      \right], 
   \end{aligned}
   \end{equation}
   where 
   \begin{equation*}
   \begin{aligned}
       r(w) &= \min\{p_1^n(w), p_2^n(w)\}, \\
       r(wa) &= \min\{p_1^{n+1}(wa), p_2^{n+1}(wa)\}.
   \end{aligned}
   \end{equation*}
\end{thm}

Theorem~\ref{thm:recursion} allows to prove that Algorithm~\ref{alg:kantorovich} efficiently computes the Kantorovich metric between $p_1^n$ and $p_2^n$. 

\begin{algorithm}[ht!]
   \caption{$\textsc{Kant}(k, m, w, n)$}\label{alg:kantorovich}
   \begin{algorithmic}
      \For{$i = 1, \dots, |\mathcal{A}|$}
         \State Compute $p_1^n(wa_i)$ and $p_2^n(wa_i)$ (see Remark~\ref{rem:prob_complexity}) 
         \State $r_i \gets \min \{ p_1^n(wa_i), p_2^n(wa_i) \}$
      \EndFor
      \State $\textsc{res} = 2^{-(k+1)}( m - \sum_{i = 1, \dots, |\mathcal{A}|} r_i)$
      \If{$k + 1 = n$}
         \State \Return \textsc{res}
      \EndIf

      \For{$i = 1, \dots, |\mathcal{A}|$} 
         \If{$r_i \neq 0$}
            \State $\textsc{res} \gets \textsc{res} + \textsc{Kant}(k + 1, r_i, wa_i, n)$ 
         \EndIf
      \EndFor
      \Return \textsc{res}
   \end{algorithmic}
\end{algorithm}

\begin{cor} \label{cor:algo}
   Let $\textnormal{\textsc{Kant}}$ be the algorithm described in Algorithm~\ref{alg:kantorovich}, then 
   \begin{equation} \label{eq:algo_term}
      K(p^n_1, p_2^n) = \textnormal{\textsc{Kant}}(0, 1, \Lambda, n).
   \end{equation}
   Moreover $\textnormal{\textsc{Kant}}$ terminates in $\mathcal{O}(|\mathcal{S}|^2|\mathcal{A}|^{n+1})$ operations.
\end{cor}

\subsection{A metric between Markov chains}
Let $\Sigma_1$ and $\Sigma_2$ be two Markov chains defined on the same alphabet $\mathcal{A}$. We define a metric between them as 
\begin{equation*} 
   d(\Sigma_1, \Sigma_2) = \lim_{n \to \infty} K(p_1^n, p_2^n), 
\end{equation*}
where $p_1^n$ and $p_2^n$ are the distributions on $\mathcal{A}^n$ induced by each Markov chain on $n$-long label sequences.
\begin{rem} \label{rem:discounted_measure}
   The Cantor distance $2^{-l}$ can be interpreted as a discount factor. Therefore, the metric $d(\Sigma_1, \Sigma_2)$, if well-defined, can be interpreted as a discounted measure of the difference between the behaviors $\mathcal{B}(\Sigma_1)$ and $\mathcal{B}(\Sigma_2)$.
\end{rem}

We now prove that this metric is well-defined\footnote{That is, the function satisfies positivity, symmetry and triangle inequality.}.
\begin{thm} \label{thm:distance}
   The metric $d(\Sigma_1, \Sigma_2)$ is well-defined. Moreover, for any $n \geq 1$,
   \begin{equation*} 
      0 \leq d(\Sigma_1, \Sigma_2) - K(p_1^n, p_2^n) \leq 2^{-n}.
   \end{equation*}
\end{thm}
Theorem~\ref{thm:distance} provides a guarantee on the approximation of $d(\Sigma_1, \Sigma_2)$ that we will be able to compute. Indeed, for any $\varepsilon > 0$, if $n \geq \lceil \log_2(\varepsilon^{-1}) \rceil$, then 
$
   0 \leq d(\Sigma_1, \Sigma_2) - K(p_1^n, p_2^n) \leq \varepsilon.
$
Following Corollary~\ref{cor:algo}, for a fixed number of labels and states, this implies that an $\varepsilon$-solution can be found in $\mathcal{O}(\varepsilon^{-1})$ computational complexity. 

\section{\textsc{Application to data-driven model abstractions}}
\label{sec:abstractions}

We now show how the metric $d(\Sigma_1, \Sigma_2)$ enables an adaptive refinement procedure for dynamical systems abstraction.

\subsection{Abstractions with adaptive refinement}

In this section, we introduce a new abstraction based on adaptive refinements. Even though our approach can be generalized to stochastic systems, in this preliminary work we focus on deterministic ones, which we now define.
\begin{defn}[Dynamical system] \label{def:system}
   A \emph{dynamical system} is the $4$-tuple $S = (X, \mathcal{A}, F, H)$ that defines the relation
   \begin{equation*}
         x_{k + 1} = F(x_k), \quad
         y_k = H(x_k),
   \end{equation*} 
   where $X \subseteq \mathbb{R}^d$ is the \emph{state space}, $\mathcal{A}$ is a finite alphabet called the \emph{output space}, $F : X \to X$ is a \emph{transition function}, and $H : X \to \mathcal{A}$ is the \emph{output function}. The variables $x_k$ and $y_k$ are called the \emph{state} and the \emph{output} at time $k$.
\end{defn}

Also, in parallel to the definition of behavior of a Markov chain, we define the \emph{behavior of a dynamical system} $\mathcal{B}(S) \subseteq \mathcal{A}^*$ as follows. A sequence $w^* = (a_1, a_2, \dots) \in \mathcal{B}(S)$ if there exists $x_1, x_2, \dots \in X$ such that $x_{i+1} = F(x_i)$ and $H(x_i) = a_i$. Also, in parallel to equivalent classes \eqref{eq:equiv_markov} on Markov chains, we define equivalent classes on the continuous state space $X$. A subset of states is an \emph{equivalent class} if it satisfies the recursive relation
\begin{equation*}
   [wa]_S = \{ x \in [w]_S \, | \, H^n(x) = a \}, \quad
   [\Lambda]_S = X,
\end{equation*}
for any $w \in \mathcal{A}^n$ and $a \in \mathcal{A}$. In other words, for a given sequence $w = (a_1, \dots, a_n)$, a state $x \in [w]_S$ if $H(x) = a_1$, $H(F(x)) = a_2$, $\dots$, and $H(F^{n-1}(x)) = a_n$. In this work, we impose the following assumption on dynamical systems.
\begin{ass} \label{ass:no_manifold}
   The dynamical system $S$ as defined as Definition~\ref{def:system} is such that, for any $w \in \mathcal{A}^{n}$ and $a \in \mathcal{A}$, the following two conditions hold:
   \begin{itemize}
      \item If $\lambda_X([w]_S) = 0$, then $[w]_S = \emptyset.$
      \item If $\lambda_X([wa]_S) = \lambda_X([w]_S)$, then $[w]_S = [wa]_S$.
   \end{itemize}
\end{ass}
Informally, Assumption~\ref{ass:no_manifold} requires that any possible trajectory has a nonzero probability to be sampled.

\begin{defn}[Adaptive partitioning] \label{def:partitioning}
   Let $w_1 \in \mathcal{A}^{n_1}$, $w_2 \in \mathcal{A}^{n_2}$, $\dots$, $w_k \in \mathcal{A}^{n_k}$ be $k$ sequences of labels of different lengths. The set of sequences $\mathcal{W} = \{w_i\}_{i = 1, \dots, k}$ is an \emph{adaptive partitioning} for $S$ if 
   \begin{equation*}
      \bigcup_{w \in \mathcal{W}} [w]_S = X, \quad
      \forall i \neq j, \, [w_i]_S \cap [w_j]_S = \emptyset. 
   \end{equation*} 
\end{defn}
We now introduce an abstraction procedure based on an adaptive partitioning refinements. 
\begin{defn}[Abstraction based on adaptive refinements] \label{def:abstraction}
   Let $S = (X, \mathcal{A}, F, H)$ be a dynamical system as defined in Definition~\ref{def:system}, and let $\mathcal{W}$ be an adaptive partitioning for $S$ as defined in Definition~\ref{def:partitioning}. Then the corresponding \emph{abstraction based on adaptive refinements} is the Markov chain $\Sigma_{\mathcal{W}} = (\mathcal{S}, \mathcal{A}, P, \mu, L)$ defined as follows:
   \begin{itemize}
      \item The states are the partitions, that is $\mathcal{S} = \mathcal{W}$.
      \item $\mu_{w}$ is the Lebesgue measure of equivalent class $[w]_S$ on $X$, that is
      $
         \mu_{w} = \lambda_X([w]_S)
      $.
      \item For $w_1 = (a_1, \dots, a_{n_1})$, and $w_2 = (b_1, \dots, b_{n_2})$, let $k = \min\{n_1 - 1, n_2\}$, $w_1' = (a_2, \dots, a_{k+1})$, and $w_2' = (b_1, \dots, b_k)$. If $w_1' \neq w_2'$ or $\lambda_X([w_1]_S) = 0$, then $P_{w_1, w_2} = 0$. Else 
      \begin{equation*}
         P_{w_1, w_2} = \frac{\lambda_X([a_1w_2]_S)}{\lambda_X([w_1]_S)}.
      \end{equation*}
      \item For $w = (a_1, \dots, a_n)$, $L(w) = a_1$.
   \end{itemize}
\end{defn}
Informally, for a given adaptive partitioning $\mathcal{W}$, the abstraction $\Sigma_{\mathcal{W}}$ can be interpreted as follows. The initial probability to be in the state $w$ in the Markov chain is the proportion of $[w]_S$ in $X$, and the probability to jump from the state $w_1$ to the state $w_2$ is the proportion of $[w_1]_S$ that goes into $[w_2]_S$ given the dynamics. For any sequence $w = (a_1, \dots, a_n) \in \mathcal{A}^n$, the probability $p^n(w)$ as defined in \eqref{eq:distribution} is therefore the approximation for our abstraction of the probability that the output signal starts with the sequence $w$.

We now provide a result that gives a sufficient condition for the abstraction to have the same behavior as the original system.
\begin{prop} \label{prop:sufficient_stop}
   Given a dynamical system $S$ satisfying Assumption~\ref{ass:no_manifold}, consider abstraction $\Sigma_{\mathcal{W}}$ as per Definition~\ref{def:abstraction}. If for all $w_1, w_2 \in \mathcal{W}$, $P_{w_1, w_2} \in \{0, 1\}$, then $\mathcal{B}(\Sigma_\mathcal{W}) = \mathcal{B}(S)$.
\end{prop}

\subsection{A data-driven abstraction}

In this section, we propose a method to construct an abstraction based on adaptive refinements, from a data set comprising outputs
sampled from the dynamical model $S$. 
Given an adaptive partitioning $\mathcal{W}$, we propose to construct $\Sigma_\mathcal{W}$ using empirical probabilities (see \cite{pmlr-v211-banse23a} for more details). We make the following assumption, which considers an idealised situation where one has an infinite number of samples. In practice, one typically has access to a finite number of observations, leading to approximation errors. The techniques to study these errors are investigated, for instance, in \cite{Coppola2023}, and are left for further work in the context of this work.
\begin{ass} \label{ass:enough_samples}
   For any abstraction $\Sigma_\mathcal{W} = (\mathcal{W}, \mathcal{A}, P, \mu, L)$, the transition probabilities $P$ and the initial distribution $\mu$ are known exactly.
\end{ass}
Now we are able to use the tool investigated in Section~\ref{sec:kantorovich} to find a smart adaptive partitioning. Indeed, one can construct two abstractions $\Sigma_{\mathcal{W}_1}$ and $\Sigma_{\mathcal{W}_2}$ corresponding to two different partitionings, and efficiently compute the Kantorovich metric $d(\Sigma_{\mathcal{W}_1}, \Sigma_{\mathcal{W}_2})$ up to some accuracy $\varepsilon$ following Corollary~\ref{cor:algo}. This gives a discounted measure of the difference between $\mathcal{B}(\Sigma_{\mathcal{W}_1})$ and $\mathcal{B}(\Sigma_{\mathcal{W}_2})$ (see Remark~\ref{rem:discounted_measure}). This reasoning leads to the greedy procedure $\textsc{Refine}(S, N, \varepsilon)$ described in Algorithm~\ref{alg:adaptive}. 
\begin{algorithm}[ht!]
   \caption{$\textsc{Refine}(S, N, \varepsilon)$}\label{alg:adaptive}
   \begin{algorithmic}
      \State $\mathcal{W} \gets \{(a)\}_{a \in \mathcal{A}}$
      \State Construct $\Sigma_\mathcal{W}$ from samples of $S$
      \While{$\exists w_1, w_2 \in \mathcal{W} : P_{w_1, w_2} \in (0, 1)$}
         \If{$N = 0$}
            \State \Return $\Sigma_{\mathcal{W}}$
         \EndIf
         \For{$i = 1, \dots, |\mathcal{W}|$}
            \State $\mathcal{W}_i' \gets \mathcal{W} \setminus \{w_i\}$
            \State $\mathcal{W}_i' \gets \mathcal{W}_i' \cup \{w_ia\}_{a \in \mathcal{A}}$ 
            \State Construct $\Sigma_{\mathcal{W}'_i}$ from samples of $S$
            \State $d_i \gets d(\Sigma_\mathcal{W}, \Sigma_{\mathcal{W}'_i})$ with precision $\varepsilon$
         \EndFor
         \State $j = \arg\max_{i = 1, \dots, |\mathcal{W}|} d_i$
         \State $\mathcal{W} \gets \mathcal{W}'_j$
         \State $\Sigma_\mathcal{W} \gets \Sigma_{\mathcal{W}'_j}$
         \State $N \gets N - 1$
      \EndWhile
      \Return $\Sigma_{\mathcal{W}}$
   \end{algorithmic}
\end{algorithm}

\begin{figure}[ht!]
   \centering
   \includegraphics[width=0.8\columnwidth]{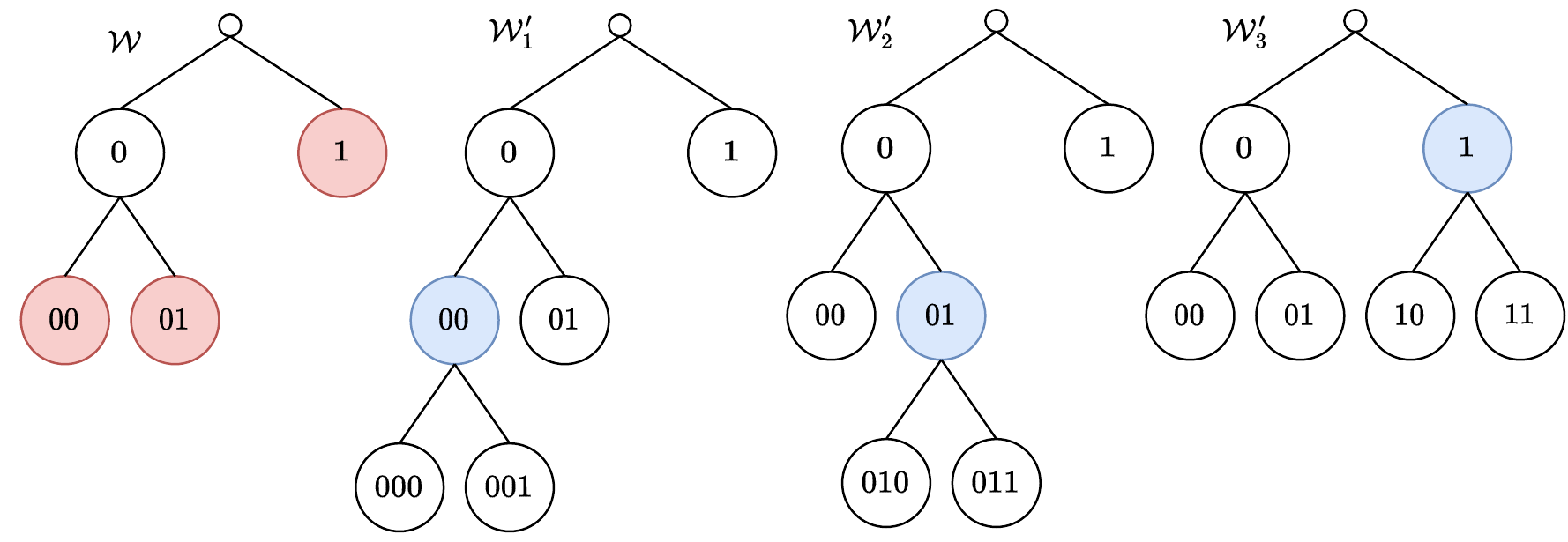}
   \caption{Illustration of the execution of Algorithm~\ref{alg:adaptive}. Consider a current partitioning $\mathcal{W} = \{00, 01, 1\}$, with the corresponding abstraction $\Sigma_\mathcal{W}$. Then the algorithm will explore the partitionings $\mathcal{W}'_1 = \{000, 001, 01, 1\}$, $\mathcal{W}'_2 = \{00, 010, 011, 1\}$ and $\mathcal{W}'_3 = \{00, 01, 10, 11\}$. For each one, it will compute $\Sigma_{\mathcal{W}'_i}$, and $d(\Sigma_{\mathcal{W}}, \Sigma_{\mathcal{W}'_i})$, and choose the one for which the distance to $\Sigma_{\mathcal{W}}$ is the largest.}
   \label{fig:execution}
\end{figure}

An interpretation of Algorithm~\ref{alg:adaptive} goes as follows. Let $\mathcal{W}$ be a coarse partitioning, and $\mathcal{W}'_1$ and $\mathcal{W}'_2$ be two more refined partitionings. If $d(\Sigma_\mathcal{W}, \Sigma_{\mathcal{W}'_1}) > d(\Sigma_\mathcal{W}, \Sigma_{\mathcal{W}'_2})$, then one could argue that it is more interesting to choose $\mathcal{W}'_1$ over $\mathcal{W}'_2$, since the discounted measure between the behaviors corresponding to the coarse partitioning and the refined partitioning is larger. Moreover, if at some point $\Sigma_\mathcal{W}$ is such that $P_{w, w'} \in \{0, 1\}$ for all $w, w' \in \mathcal{A}^n$, then one has a sufficient condition to stop the algorithm following Proposition~\ref{prop:sufficient_stop}, otherwise the algorithm stops after $N$ iterations. If $N = \infty$, then the algorithm only stops in such case. An execution step of the algorithm can be found in Figure~\ref{fig:execution}. A complexity analysis of Algorithm~\ref{alg:adaptive} can be found in Corollary~\ref{cor:algo2}.

\begin{cor} \label{cor:algo2}
   The algorithm $\textnormal{\textsc{Refine}}(S, N, \varepsilon)$ terminates in $\mathcal{O}(|\mathcal{A}|^{n+4}N^4)$ operations, with $n = \lceil\log_2(\varepsilon^{-1})\rceil$. Moreover, for $S$ satisfying Assumption~\ref{ass:no_manifold}, if $\Sigma_\mathcal{W} = \textnormal{\textsc{Refine}}(S, \infty, \varepsilon)$ terminates, then $\mathcal{B}(\Sigma_\mathcal{W}) = \mathcal{B}(S)$.
\end{cor}

\subsection{Numerical examples}
In this section, we demonstrate on an example that our greedy algorithm converges to a smart partitioning\footnote{All the code corresponding to this section can be found at \url{https://github.com/adrienbanse/KantorovichAbstraction.jl}.}, and we show how to use the proposed framework for controller design.

\begin{exmp} \label{exmp:memory}
   Consider $S = (X, \mathcal{A}, F, H)$ with $X = [0, 2] \times [0, 1]$, $\mathcal{A} = \{0, 1\}$. Let $F$ be defined as 
   \begin{equation*} 
      F(x) = \begin{cases}
         x &\textrm{if } x \in P_1 \cup P_4, \\
         (x_1/2  + 1/2, x_2 + 1/2) &\textrm{if } x \in P_2, \\
         (x_1 - 1/2, x_2) &\textrm{if } x \in P_3, \\
         (2x_1 + 1, 4x_2 - 3/4) &\textrm{else}, 
      \end{cases} 
   \end{equation*}
   where $P_i$ are depicted in Figure~\ref{fig:system}, and $H(x) = 0$ if $x \in P_1$, else $H(x) = 1$.
   An illustration and interpretation of $S$ is given in Figure~\ref{fig:system}.
   \begin{figure}[ht!]
      \centering
      \includegraphics[width = 0.5\columnwidth]{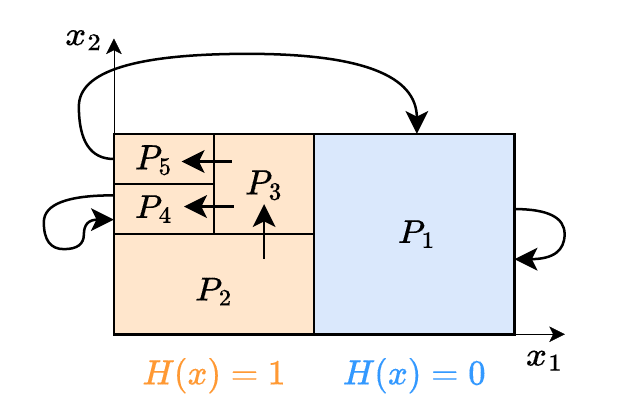}
      \caption{Illustration and description of the transition function $F$ of Example~\ref{exmp:memory}. $F$ has to be understood in the following way: $P_1$ is mapped to itself, $P_2$ is mapped to $P_3$, the lower half of $P_3$ is mapped to $P_4$, the upper half of $P_3$ is mapped to $P_5$, $P_5$ is mapped to $P_1$, and $P_4$ to itself.}
      \label{fig:system}
   \end{figure}
\end{exmp}

\begin{figure}[ht!]
   \centering
   \includegraphics[width = 0.5\columnwidth]{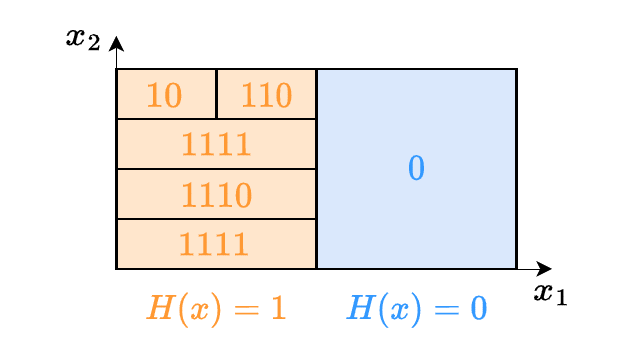}
   \caption{Illustration of the last partitioning $\mathcal{W}$ given by Algorithm~\ref{alg:adaptive} for Example~\ref{exmp:memory}.}
   \label{fig:partition}
\end{figure}

The result of the algorithm applied to Example~\ref{exmp:memory} at all iterations $k$ is depicted in Table~\ref{tab:refine}, and the final partitioning is illustrated in Figure~\ref{fig:partition}. Observe that the generated partitioning aligns well with the dynamics, and that our algorithm generates an emerging structure which is not trivial. The algorithm stops at the third iteration since the obtained abstraction is such that $P_{w, w'} \in \{0, 1\}$, which is a stopping criterion following Proposition~\ref{prop:sufficient_stop}, and has much less states than the brute force approach of \cite{pmlr-v211-banse23a}. 

\begin{table}[ht!]
   \begin{tabular}{cccc}
      & $\mathcal{W}$ & $d(\Sigma_\mathcal{W}, \Sigma_{\mathcal{W}'_j})$ & $P_{w, w'} \in \{0, 1\}$\\
      \hline
      $k=0$ & $\{0, \textbf{1}\}$              & $0.0015$ & No  \\
   
      $k=1$ & $\{0, 10, \textbf{11}\}$              & $0.0059$ & No  \\
      $k=2$ & $\{0, 10, 110, \textbf{111}\}$        & $0.0039$ & No  \\
      $k=3$ & $\{0, 10, 110, 1110, 1111\}$ & - & \textbf{Yes}
   \end{tabular}
   \caption{Results of Algorithm~\ref{alg:adaptive} for Example~\ref{exmp:memory}. For each iteration $k$, the current model is the abstraction corresponding to $\Sigma_{\mathcal{W}}$, and the chosen model is $\Sigma_{\mathcal{W}'_j}$, with the largest distance $d(\Sigma_{\mathcal{W}}, \Sigma_{\mathcal{W}_j'})$. With $P$ the transition matrix of the current model, if for all $w, w' \in \mathcal{A}^n$, $P_{w, w'} \in \{0, 1\}$, the algorithm stops.}
   \label{tab:refine}
\end{table}

We further demonstrate the quality of the obtained abstractions by designing a controller for a similar dynamical system.
\begin{exmp} \label{exmp:memory_control}
   Consider the dynamical system $S$ as described in Example~\ref{exmp:memory}, except that the dynamics is controlled as follows:
   \begin{equation*}
      \tilde{x}_k = x_k + \begin{pmatrix}
         0 \\ 1
      \end{pmatrix}u_k, \quad
      x_{k+1} = F(\tilde{x}_k)
   \end{equation*}
   where $u_k = K(x_k) \in \{0, 1/4, 1/2\}$ is an input to the system. Consider the reward $r(x) = 1$ if $H(x) = 0$, else $r(x) = 0$, and a discounted expected reward maximization objective, that is 
   \begin{equation} \label{eq:reward_system}
      \max_{K} \mathbb{E}_{x_1 \sim \mu(X)} \sum_{k} \gamma^k r(x_k), 
   \end{equation}
   where $\mu(X)$ is the uniform distribution on $X$, and $\gamma = 0.95$ is a discount factor.
\end{exmp}

To solve this optimal control problem, we will use the abstractions constructed by Algorithm~\ref{alg:adaptive}. For each partitioning $\mathcal{W}$ in Table~\ref{tab:refine}, we will construct the abstraction $\Sigma_\mathcal{W}^{u}$ corresponding to the actions given in Example~\ref{exmp:memory_control}, that is $u = 0$, $u = 1/4$ and $u = 1/2$. We will then solve a Markov Decision Process (or MDP for short, see \cite{Sutton2018} for an introduction) maximizing the expected reward of the MDP. For this, we used the implementation of the value iteration algorithm implemented in the $\texttt{POMDPs.jl}$ Julia package \cite{egorov2017pomdps}. Now, let $P(s)$ be the optimal policy for the state $s$, we design the controller for the system \ref{exmp:memory_control} as follows: for $x_k \in [w]_S$, then 
\begin{equation} \label{eq:controller}
    u_k = K(x_k) = P(w).
\end{equation}
 For the different abstractions found by Algorithm~\ref{alg:adaptive}, the corresponding expected rewards \eqref{eq:reward_system} for the original system controlled by \eqref{eq:controller} are given in Table~\ref{tab:MDP}. One can see that the expected reward increases as our algorithm refines the state-space.
\begin{table}[ht!]
\centering
\begin{tabular}{ccc}
Iteration & Controller \eqref{eq:controller} & \textbf{Expected reward \eqref{eq:reward_system}} \\ \hline
$k=0$& 
$
    K(x) = \begin{cases}
        0 &\textrm{if } x \in [0]_S\\
        0 &\textrm{if } x \in [1]_S
    \end{cases}
$
& $\textbf{14.4784}$         \\ 
$k=1$ & 
$
    K(x) = \begin{cases}
        0 &\textrm{if } x \in [0]_S\\
        0 &\textrm{if } x \in [10]_S\\
        1/4 &\textrm{if } x \in [11]_S
    \end{cases}
$
& $\textbf{18.8726}$         \\
$k=2$& 
$
    K(x) = \begin{cases}
        0 &\textrm{if } x \in [0]_S\\
        0 &\textrm{if } x \in [10]_S\\
        0 &\textrm{if } x \in [110]_S\\
        1/4 &\textrm{if } x \in [111]_S
    \end{cases}
$
& $\textbf{19.0311}$        \\
$k=3$& 
$
    K(x) = \begin{cases}
        0 &\textrm{if } x \in [0]_S\\
        0 &\textrm{if } x \in [10]_S\\
        0 &\textrm{if } x \in [110]_S\\
        1/2 &\textrm{if } x \in [1110]_S\\
        1/4 &\textrm{if } x \in [1111]_S
    \end{cases}
$
& $\textbf{19.1022}$        
\end{tabular}
\caption{Expected rewards \eqref{eq:reward_system} for the Example~\ref{exmp:memory_control} controlled by \eqref{eq:controller}. The iterations $k$ correspond to the iterations of Algorithm~\ref{alg:adaptive} represented in Table~\ref{tab:refine}. The optimal policy is found by solving MDPs corresponding to the three possible actions $u_k \in \{0, 1/4, 1/2\}$, and the expected reward \eqref{eq:reward_system} is approximated by sampling $5000$ trajectories of length $1000$. One can observe that the expected reward increases.}
\label{tab:MDP}
\end{table}

\section{\textsc{Conclusion and further research}}

Inspired by a recent interest in developing data-driven abstractions of dynamical systems, we proposed a state refinement procedure that relies on a Kantorovich metric between Markov chains. We leverage the Cantor distance in the space of behaviours of the generated abstraction and use it to define the proposed Kantorovich metric. A key feature of our approach is a greedy strategy to perform state refinement that leads to an adaptive and smart partition of the state space. We show promising results in some control problems.


As further research, we plan to design a smart stopping criterion for our refinement procedure. We would also like to investigate convergence properties of our method, in the same fashion as in \cite[Theorem~8]{pmlr-v211-banse23a}.


\section*{\textsc{Acknowledgment}}
We thank Prof. Franck van Breugel and Prof. Prakash Panangaden for their insightful comments and the interesting conversations about this work.

\bibliographystyle{IEEEtran}
\bibliography{cdc23.bib}

\IfCustom[\proofs]{
   \appendix

   \subsection{Proof of Lemma~\ref{lem:min}}
   We first prove that $\pi^n(w, w) \leq \min(p_1^n(w), p_2^n(w))$. Constraints \eqref{eq:linear_constraints} imply that, for all $w \in \mathcal{A}^n$,
   \begin{equation}
   \begin{aligned}
      p_1^n(w) = \pi^n(w, w) + \sum_{\substack{w' \in \mathcal{A}^n\\w' \neq w}} \pi^n(w, w') \geq \pi^n(w, w), \\
      p_2^n(w) = \pi^n(w, w) + \sum_{\substack{w' \in \mathcal{A}^n\\w' \neq w}} \pi^n(w', w) \geq \pi^n(w, w), 
   \end{aligned}
   \end{equation}
   which imply that $\pi^n(w, w) \leq \min\{p_1^n(w), p_2^n(w)\}$. Now we prove that $\pi^n(w, w) \geq \min\{p_1^n(w), p_2^n(w)\}$. Consider an optimal solution $\pi^n$ to the problem~\eqref{eq:kantorovich} such that
   \begin{equation}
      \pi^n(w, w) = \min\{p_1^n(w), p_2^n(w)\} - \varepsilon, 
   \end{equation} 
   for some $w \in \mathcal{A}^n$ and $\varepsilon > 0$. Assume w.l.o.g. that $\min\{p_1(w), p_2(w)\} = p_1(w)$. Therefore, constraints~\eqref{eq:linear_constraints} imply that 
   \begin{enumerate}
      \item there exists $w' \neq w$, such that $\pi^n(w, w') = \varepsilon'$ for some $\varepsilon' \in (0, \varepsilon]$, and
      \item there exists $w'' \neq w$ such that $\pi^n(w'', w) = \varepsilon''$ for some $\varepsilon'' \in (0, \varepsilon]$.
   \end{enumerate}
   Let $K(p_1^n, p_2^n)$ denote the Kantorovich metric corresponding to such $\pi^n$. Now assume w.l.o.g. that $\varepsilon' \leq \varepsilon''$. Consider then $(\pi^n)'$ such that $(\pi^n)'(w_1, w_2) = \pi^n(w_1, w_2)$ for all $w_1, w_2 \in \mathcal{A}^n$ except 
   \begin{enumerate}
      \item $(\pi^n)'(w, w) = \pi^n(w, w) + \varepsilon'$, 
      \item $(\pi^n)'(w, w') = \pi^n(w, w') - \varepsilon'$, 
      \item $(\pi^n)'(w'', w') = \pi^n(w'', w') + \varepsilon'$, and
      \item $(\pi^n)'(w'', w) = \pi^n(w, w) - \varepsilon'$.
   \end{enumerate}
   The joint distribution $(\pi^n)'$ is feasible since it still satisfies the constraints~\eqref{eq:linear_constraints}. Now let $K'(p_1^n, p_2^n)$ denote the solution corresponding to such $(\pi^n)'$, we have that
   \begin{equation}
   \begin{aligned}
      &K'(p_1^n, p_2^n) = K(p_1^n, p_2^n) \\
      &\quad - \varepsilon'\left[d_B(w, w') + d_B(w', w'') - d_B(w', w'')\right].
   \end{aligned}
   \end{equation}
   Since the Baire's distance $d_B$ as defined in Definiton~\ref{def:baire} satisfies triangular inequality, we have that $K'(p_1^n, p_2^n) < K(p_1^n, p_2^n)$, which is a contradiction.
   
   \subsection{Proof of Lemma~\ref{lem:raph}}
   For some $w \in \mathcal{A}^n$, assume w.l.o.g. that $p_1^n(w) > p_2^n(w)$. First, by feasibility conditions, 
   \begin{equation}
      \sum_{\substack{w' \in \mathcal{A}^n\\w' \neq w}} \sum_{a_1, a_2 \in \mathcal{A}} \pi^{n+1}(wa_1, w'a_2) \geq p_1^n(w) - p_2^n(w). 
   \end{equation}
   Now, we proceed similarly as for Lemma~\ref{lem:min}. Suppose by contradiction that 
   \begin{equation}
      \sum_{\substack{w' \in \mathcal{A}^n\\w' \neq w}} \sum_{a_1, a_2 \in \mathcal{A}} \pi^{n+1}(wa_1, w'a_2) > p_1^n(w) - p_2^n(w). 
   \end{equation}
   Then there exists $w' \neq w \in \mathcal{A}^n$, and $a_1, a_2 \in \mathcal{A}$ such that $\pi^{n+1}(w'a_1, wa_2) = \varepsilon' > 0$. There also exists $w'' \in \mathcal{A}^n$ such that $w'' \neq w$ and $w'' \neq w'$, and $a_3, a_4 \in \mathcal{A}$ such that $\pi^{n+1}(wa_3, w''a_4) = \varepsilon'' > 0$. Asumme w.l.o.g. that $\varepsilon' \leq \varepsilon''$, and consider a solution $(\pi^{n+1})'$ such that $(\pi^{n+1})' = \pi^{n+1}$, except for 
   \begin{enumerate}
      \item $(\pi^{n+1})'(wa_3, w''a_4) = \pi^{n+1}(wa_3, w''a_4) - \varepsilon'$, 
      \item $(\pi^{n+1})'(w'a_1, wa_2) = \pi^{n+1}(w'a_1, wa_2) - \varepsilon'$, 
      \item $(\pi^{n+1})'(w'a_1, w''a_4) = \pi^{n+1}(w'a_1, w''a_4) + \varepsilon'$, and
      \item $(\pi^{n+1})'(wa_3, wa_2) = \pi^{n+1}(wa_3, wa_2) + \varepsilon'$.
   \end{enumerate} 
   The joint distribution $(\pi^{n+1})'$ is feasible since it still satisfies the constraints~\eqref{eq:linear_constraints}. Note that, since the Baire's distance satisfies the strong triangular inequality (see Remark~\ref{rem:ultra}), 
   \begin{equation}
   \begin{aligned}
      d_B(w'a_1, w''a_4) &\leq \max\{d_B(w'a_1, wa_2), d_B(wa_2, w''a_4)\}\\
      &= \max\{d_B(w'a_1, wa_2), d_B(wa_3, w''a_4)\}.
   \end{aligned}
   \end{equation}
   Moreover, $d_B(wa_3, wa_2) = 2^{-(n+1)}$. Now let $K'(p_1^{n+1}, p_2^{n+1})$ denote the solution corresponding to such $(\pi^{n+1})'$, we have that $K(p_1^{n+1}, p_2^{n+1}) - K'(p_1^{n+1}, p_2^{n+1})$ is
   \begin{equation}
   \begin{aligned}
      &- \varepsilon'\left[
         \begin{array}{ll}
            +& d_B(w'a_1, wa_2) \\
            +& d_B(wa_3, w''a_4) \\
            -&d_B(w'a_1, w''a_4)\\ 
            -& 2^{-(n+1)}
         \end{array}
      \right]\\
      & \leq - \varepsilon'\left[
         \begin{array}{ll}
            +& d_B(w'a_1, wa_2) + d_B(wa_3, w''a_4) \\
            -&\max\{d_B(w'a_1, wa_2), d_B(wa_3, w''a_4)\}\\ 
            -& 2^{-(n+1)}
         \end{array}
      \right] \\
      &\leq - \varepsilon'[2^{-n} - 2^{-(n+1)}] \\
      &\leq 0, 
   \end{aligned}
   \end{equation}
   which contradicts the fact that $\pi^{n+1}$ is optimal.
   
   \subsection{Proof of Theorem~\ref{thm:recursion}}
   For the sake of clarity, we note $K^n = K(p_1^n, p_2^n)$. We first prove that the right hand side of \eqref{eq:iterative_prop} is a lower bound for $K^{n+1}$. First we note that $K^{n+1}$ is equal to
   \begin{equation}
   \begin{aligned}
      &\sum_{w_1, w_2 \in \mathcal{A}^n} \sum_{a_1, a_2 \in \mathcal{A}} d_B(w_1a_1, w_2a_2) \pi^{n+1}(w_1a_1, w_2a_2) \\
      =& \sum_{\substack{w_1, w_2 \in \mathcal{A}^n\\w_1 \neq w_2}} d_B(w_1, w_2) \sum_{a_1, a_2 \in \mathcal{A}} \pi^{n+1}(w_1a_1, w_2a_2) \\
      &+ 2^{-(n+1)} \sum_{w \in \mathcal{A}^n} \sum_{\substack{a_1, a_2 \in \mathcal{A}\\a_1 \neq a_2}} \pi^{n+1}(wa_1, wa_2) \\
      &:= C_1 + C_2.
   \end{aligned}
   \end{equation}
   We first prove that $C_1 \geq K^n$. To do this, let $\mu^n : \mathcal{A}^n \times \mathcal{A}^n \to [0, 1]$ be defined as 
   \begin{equation}
      \mu^n(w_1, w_2) = \sum_{a_1, a_2 \in \mathcal{A}} \pi^{n+1}(w_1a_1, w_2a_2).
   \end{equation}
   We show that $\mu^n$ satisfies the constraints \eqref{eq:linear_constraints}. Indeed $\mu^n(w_1, w_2) \geq 0$, 
   \begin{equation}
   \begin{aligned}
      \sum_{w_2 \in \mathcal{A}^n} \mu(w_1, w_2) &= \sum_{w_2 \in \mathcal{A}^n} \sum_{a_1, a_2 \in \mathcal{A}} \pi^{n+1}(w_1a_1, w_2a_2)\\
      &= \sum_{a_1} p_1^{n+1}(w_1a_1)\\
      &= p_1^n(w_1), 
   \end{aligned}
   \end{equation}
   and similarly for the third condition in \eqref{eq:linear_constraints}. This implies that $\mu^n$ is a coupling, thereby a feasible solution of $\eqref{eq:kantorovich}$. This yields 
   \begin{equation}
      K^n \leq \sum_{a_1, a_2 \in \mathcal{A}} d_B(w_1, w_2) \mu^n(w_1, w_2) = C_1.
   \end{equation} 
   Now we show that, for all $w \in \mathcal{A}^n$, 
   \begin{equation}
      \sum_{\substack{a_1, a_2 \in \mathcal{A}\\a_1 \neq a_2}} \pi^{n+1}(wa_1, wa_2) = r(w) - \sum_{a \in \mathcal{A}} r(wa), 
   \end{equation}
   which implies that 
   \begin{equation}
      C_2 = 2^{-(n+1)} \sum_{w \in \mathcal{A}^n} \left[ \sum_{w \in \mathcal{A}^n} r(w) - \sum_{a \in \mathcal{A}} r(wa) \right].
   \end{equation}
   We prove the claim. Assume w.l.o.g. that $w$ is such that $p_1^n(w) > p_2^n(w)$, then 
   \begin{equation}
   \begin{aligned}
      &\sum_{a_1, a_2 \in \mathcal{A}} \pi^{n+1}(wa_1, wa_2) \\
      & = \sum_{a_1 \in \mathcal{A}} 
         \sum_{w' \in \mathcal{A}^n} \sum_{a_2 \in \mathcal{A}} 
         \pi^{n+1}(wa_1, w'a_2) \\
      & \quad \quad \quad \quad - \sum_{\substack{w' \in \mathcal{A}^n\\w' \neq w}}
      \sum_{a_2 \in \mathcal{A}} 
      \pi^{n+1}(wa_1, w'a_2) \\
      &= \sum_{a_1 \in \mathcal{A}} p_1^n(wa_1) - \sum_{\substack{w' \in \mathcal{A}^n\\w' \neq w}} \sum_{a_1, a_2 \in \mathcal{A}} 
      \pi^{n+1}(wa_1, w'a_2) \\
   \end{aligned}  
   \end{equation}
   Following Lemma~\ref{lem:raph}, this is equal to 
   \begin{equation}
      p_1^n(w) - \left(p_1^n(w) - p_2^n(w)\right) = r(w).
   \end{equation}
   And the following holds: 
   \begin{equation}
   \begin{aligned}
      &\sum_{\substack{a_1, a_2 \in \mathcal{A}\\a_1 \neq a_2}} \pi^{n+1}(wa_1, wa_2) \\
      =& \sum_{\substack{a_1 \neq a_2 \in \mathcal{A}\\a_1 \neq a_2}} \pi^{n+1}(wa_1, wa_2) - \sum_{a \in \mathcal{A}} \pi^{n+1}(wa_1, wa_2) \\
      =&\, r(w) - \sum_{a \in \mathcal{A}} r(wa)
   \end{aligned}
   \end{equation}
   by Lemma~\ref{lem:min}. This concludes that the right hand side of \eqref{eq:iterative_prop} is a lower bound for $K^{n+1}$. 
   
   Now, to provide an upper bound, we will show that we can construct a feasible $n + 1$ solution feasible $\mu^{n+1}$ such that 
   \begin{equation}
   \begin{aligned}
      &\sum_{w_1, w_2 \in \mathcal{A}^{n+1}} d_B(w_1, w_2)\mu^{n+1}(w_1, w_2) \\
      &\quad \quad \quad = K^n + \sum_{w \in \mathcal{A}^n} \left[
         r(w) - \sum_{a \in \mathcal{A}} r(wa)
      \right].
   \end{aligned}
   \end{equation}
   Consider $\pi^n$, an optimal solution at step $n$. We will construct $\mu^{n+1}$ in the following greedy way. Initialize $\mu^{n+1}$ with only zero elements, and for all $w \in \mathcal{A}^{n}$, $a \in \mathcal{A}$, we initialize $\delta(wa) = 0$. We start by updating the blocks $\mu^{n+1}(w_1a_1, w_2a_2)$ where $w_1 \neq w_2$. For all $w$ such that $p_1^{n}(w) > p_2^{n}(w)$, for all $a \in \mathcal{A}$ such that $p_1^{n+1}(wa) > p_2^{n+1}(wa)$, do the following.
   \begin{enumerate}
      \item Let $\tilde{\delta}(wa) = p_1^{n+1}(wa) - p_2^{n+1}(wa)$.\\
      If $\sum_{a' \neq a} \delta(wa') + \tilde{\delta}(wa) > p_1^{n}(w) - p_2^{n}(w)$, let $\delta(wa) = (p_1^{n}(w) - p_2^{n}(w)) - \sum_{a' \neq a} \delta(wa')$.\\
      Else let $\delta(wa) = \tilde{\delta}(wa)$.
      \item \label{item:current} Find a $w' \neq w$ such that
      \begin{equation}
         \pi^n(w, w') > \sum_{a_1, a_2 \in \mathcal{A}} \mu^{n+1}(wa_1, w'a_2). 
      \end{equation}
      Now, for any $a' \in \mathcal{A}$, let 
      \begin{equation}
      \begin{aligned}
         &\psi(a') = p_2^{n+1}(w'a') - p_1^{n+1}(w'a')  \\
         & - \sum_{\substack{w'' \in \mathcal{A}^n\\w'' \neq w'}} \sum_{a_1 \in \mathcal{A}} \mu^{n+1}(w''a_1, w'a')
      \end{aligned}
      \end{equation}
      Then, find $a' \in \mathcal{A}$ such that $\psi(a') > 0$. \\
      Now, if $\delta(wa) > \psi(a')$, then: 
      \begin{itemize}
         \item Update $\mu(wa, w'a') \gets \psi(a')$
         \item Update $\delta(wa) \gets \delta(wa) - \psi(a')$
         \item Return to \ref{item:current}.
      \end{itemize}
      Else, update $\mu(wa, w'a') \gets \delta(wa)$.
   \end{enumerate}
   We claim that, in the procedure above, there always exists such a $w'$ for a given $wa$. Otherwise, 
   \begin{equation}
      \sum_{w' \neq w}\pi^n(w, w') < p_1^n(w) - p_2^n(w),  
   \end{equation}
   which is impossible by Lemma~\ref{lem:min}. Also, we claim that there also always exists such $a'$ for a given $wa$ and $w'$. Otherwise, for all $a' \in \mathcal{A}$, 
   \begin{equation}
   \begin{aligned}
      &\sum_{a' \in \mathcal{A}} \sum_{\substack{w'' \in \mathcal{A}^n\\w'' \neq w'}} \sum_{a_1 \in \mathcal{A}} \mu^{n+1}(w''a_1, w'a') \\
      &\quad = \sum_{a' \in \mathcal{A}} p_2^{n+1}(w'a') - p_1^{n+1}(w'a'), 
   \end{aligned}
   \end{equation}
   which means by construction that 
   \begin{equation}
   \begin{aligned}
      &\sum_{a' \in \mathcal{A}} \sum_{\substack{w'' \in \mathcal{A}^n\\w'' \neq w'}} \sum_{a_1 \in \mathcal{A}} \pi^{n+1}(w''a_1, w'a') \\
      &\quad > \sum_{a' \in \mathcal{A}} p_2^{n+1}(w'a') - p_1^{n+1}(w'a'), 
   \end{aligned}
   \end{equation}
   which is $p_2^{n}(w) - p_1^{n}(w) > p_2^{n}(w) - p_1^{n}(w)$ by Lemma~\ref{lem:raph}. Moreover, by construction we have that, for all $w \neq w'$,
   \begin{equation} \label{eq:non_diag_blocks}
      \pi^{n}(w, w') = \sum_{a_1, a_2 \in \mathcal{A}^n} \mu^{n+1}(wa_1, w'a_2).
   \end{equation}
   
   Now, we construct the diagonal blocks $\mu^{n+1}(wa_1, wa_2)$. For each $w \in \mathcal{A}^n$ and $a \in \mathcal{A}$, let 
   \begin{equation} \label{eq:fake_constraints}
   \begin{gathered}
      \tilde{p}_1^{n+1}(wa) = p_1^{n+1}(wa) - \sum_{w' \neq w} \sum_{a \in \mathcal{A}} \mu^{n+1}(wa, w'a'),\\
      \tilde{p}_2^{n+1}(wa) = {p}_2^{n+1}(wa) - \sum_{w' \neq w} \sum_{a \in \mathcal{A}} \mu^{n+1}(w'a', wa).
   \end{gathered}
   \end{equation}
   Now, for a given $w$, let us solve the following balanced optimal transport problem: 
   \begin{equation}
   \begin{aligned}
      &\inf_{\mu^{n+1}} 2^{-(n+1)}\sum_{\substack{a_1, a_2 \in \mathcal{A}\\a_1 \neq a_2}} \mu^{n+1}(wa_1, wa_2) \\
      \textrm{s.t. }&\forall a_1 \in \mathcal{A}: \, \sum_{a_2} \mu^{n+1}(wa_1, wa_2) = \tilde{p}_1^{n+1}(wa_1), \\
      &\forall a_2 \in \mathcal{A}: \, \sum_{a_1} \mu^{n+1}(wa_1, wa_2) = \tilde{p}_2^{n+1}(wa_2).
   \end{aligned}
   \end{equation}
   Following the definition of $\tilde{p}$ and $\tilde{q}$, this is a balanced optimal transport whose trivial solution is given by 
   \begin{equation} \label{eq:trivial_balanced}
      2^{-(n+1)} \left(r(w) - \sum_{a \in \mathcal{A}} r(wa)\right).
   \end{equation}
   
   Now we conclude the proof. By \eqref{eq:non_diag_blocks} and \eqref{eq:fake_constraints}, $\mu^{n+1}$ is a coupling of $p_1^{n+1}$ and $p_2^{n+1}$. Indeed it is positive, and for any $w_1 \in \mathcal{A}^n$ and $a_1 \in \mathcal{A}$, 
   \begin{equation}
   \begin{aligned}
      &\sum_{w_2 \in \mathcal{A}^n} \sum_{a_2 \in \mathcal{A}} \mu^{n+1}(w_1a_1, w_2a_2) \\
      = &\sum_{a_2 \in \mathcal{A}} \mu^{n+1}(w_1a_1, w_1a_2) + \sum_{\substack{w_2 \in \mathcal{A}^n\\w_2 \neq w_1}} \sum_{a_2 \in \mathcal{A}} \mu^{n+1}(w_1a_1, w_2a_2) \\
      = \, &\tilde{p}_1^{n+1}(w_1a_1) + \left(p_1^{n+1}(w_1a_1) - \tilde{p}_1^{n+1}(w_1a_1)\right)\\
      = \, &p_1^{n+1}(w_1a_1), 
   \end{aligned}
   \end{equation}
   and similarly for $p_2^{n+1}$. Finally, 
   \begin{equation}
   \begin{aligned}
      &\sum_{w_1, w_2 \in \mathcal{A}^n} \sum_{a_1, a_2 \in \mathcal{A}} d_B(w_1a_1, w_2a_2) \mu^{n+1}(w_1a_1, w_2a_2) \\
      =& \sum_{w_1 \neq w_2} d_B(w_1, w_2) \sum_{a_1, a_2} \mu^{n+1}(w_1a_1, w_2a_2) \\
      +& 2^{-(n+1)} \sum_{w} \sum_{\substack{{a_1, a_2}\\a_1 \neq a_2}} \mu^{n+1}(wa_1, wa_2).
   \end{aligned}
   \end{equation}
   By \eqref{eq:non_diag_blocks}, the first term is $K^n$, and by \eqref{eq:trivial_balanced}, the second term is 
   \begin{equation}
      \sum_{w \in \mathcal{A}^n} \left[r(w) - \sum_{a \in \mathcal{A}} r(wa) \right].
   \end{equation}
   This provides an upper bound on $K^{n+1}$, and the proof is completed. 
   
   \subsection{Proof of Corollary~\ref{cor:algo}}
   We will prove that \eqref{eq:algo_term} holds by induction on the level of the execution tree of Algorithm~\ref{alg:kantorovich}. Let us prove that case $n = 1$. 
   The constraints \eqref{eq:linear_constraints} imply that 
   \begin{equation*}
      \sum_{a_1, a_2 \in \mathcal{A}} \pi^1((a_1), (a_2)) = 1.
   \end{equation*}   
   Therefore,
   \begin{equation} \label{eq:K_1}
   \begin{aligned}
      K(p^1_1, p^1_2) &= 2^{-1}
      \sum_{\substack{a_1, a_2 \in \mathcal{A}\\a_1 \neq a_2}} \pi^1((a_1), (a_2)) \\
      &= 2^{-1} \left[
         1 - \sum_{a \in \mathcal{A}} \pi^1((a), (a))
      \right].
   \end{aligned}
   \end{equation}
      Moreover, for all $a \in \mathcal{A}$, the solution of $\pi^{n}((a), (a)) = r(wa)$ by Lemma~\ref{lem:min}. Therefore \eqref{eq:K_1} is the result of $\textsc{Kant}(0, 1, \Lambda, 1)$. Now, assume that \eqref{eq:algo_term} holds for $n$. By Theorem~\ref{thm:recursion}, 
   \begin{equation} \label{eq:algo_recursion}
   \begin{aligned}
      &K(p_1^{n+1}, p_2^{n+1}) = \textsc{Kant}(0, 1, \Lambda, n) \\
      &\quad \quad+
      2^{-(n+1)}\sum_{w \in \mathcal{A}^{n}}
      \left[
         r(w)
         -
         \sum_{a \in \mathcal{A}}
         r(wa)
      \right].
   \end{aligned}
   \end{equation}
   Following the notations of Algorithm~\ref{alg:kantorovich}, let $m^w = r(w)$, and let $r_i^w = r(wa_i)$. One can re-write \eqref{eq:algo_recursion} as 
   \begin{equation*}
   \begin{aligned}
      &K(p_1^{n+1}, p_2^{n+1}) = \textsc{Kant}(0, 1, \Lambda, n)\\
      &\quad\quad+
      \sum_{w \in \mathcal{A}^{n}}
      2^{-(n+1)}
      \left[
         m^{w}
         -
         \sum_{i = 1, \dots, |\mathcal{A}|}
         r_i^{w}
      \right].
   \end{aligned}
   \end{equation*}
   One can recognize $\textsc{Kant}(0, 1, \Lambda, n+1)$ in the right hand side of the equation above, which conludes the proof of \eqref{eq:algo_term}.
   
   In terms of computational complexity, the bottleneck of Algorithm~\ref{alg:kantorovich} is the computation of $p_1^n$ and $p_2^n$ at each node of the execution tree. Following Remark~\ref{rem:prob_complexity}, this can be done in $\mathcal{O}(|\mathcal{S}|^2)$ operations. Since there are $\mathcal{O}(|\mathcal{A}|^{n+1}|)$ nodes in the execution tree, the total number of operations is $\mathcal{O}(|\mathcal{S}|^2|\mathcal{A}|^{n+1})$.
   
   \subsection{Proof of Theorem~\ref{thm:distance}} \label{app:proofthm2}
   Let $K_n = K(p_1^n, p_2^n)$. First, we prove that, for all $n \geq 1$,
   \begin{equation} \label{eq:sandwich}
      0 \leq K_{n+1} - K_n \leq 2^{-(n+1)}.
   \end{equation}
   Following Theorem~\ref{thm:recursion}, it suffices to show that
   \begin{equation} \label{eq:diff_0_1}
      0 \leq \sum_{w \in \mathcal{A}^n} 
      \left[r(w) - \sum_{a \in \mathcal{A}} r(wa)\right] \leq 1, 
   \end{equation}
      By the law of total probability, we have that 
      \begin{equation*}
         p_1^{n}(w) = \sum_{a \in \mathcal{A}} p_1^{n+1}(wa), 
      \end{equation*}
      and similarly for $p_2^n(w)$. Hence 
      \begin{equation*}
         0 \leq r(w) - \sum_{a \in \mathcal{A}} r(wa) \leq r(w), 
      \end{equation*}
      which shows that \eqref{eq:diff_0_1} holds. Now, notice that \eqref{eq:sandwich} implies that the sequence $(K_n)_{n \geq 1}$ is monotone, and bounded since
   \begin{equation*}
      \begin{aligned}
         \lim_{n \to \infty} K_n &\leq K_1 + \sum_{n \geq 1} (K_{n+1} - K_n) \\
         &\leq 2^{-1} + \sum_{n \geq 1} 2^{-(n+1)} \\
         &= 1.
      \end{aligned}
   \end{equation*}
   By the monotone convergence theorem, the limit exists and is equal to 
   \begin{equation*}
      \lim_{n \to \infty} K_n = \sup_{n \geq 1} K_n.
   \end{equation*}
   Moreover, since $K_n$ is a distance for every $n \geq 1$, and the limit exists, we have that $\lim_{n \to \infty} K_n$ is also a distance. Finally, by \eqref{eq:sandwich},
   \begin{equation*}
      \begin{aligned}
      \lim_{n \to \infty} K_n - K_p &= \left[K_1 + \sum_{n = 1}^{\infty} K_{n+1} - K_n\right] \\
      &\quad\quad\quad\quad\quad - \left[K_1 + \sum_{n = 1}^{p-1} K_{n+1} - K_n\right] \\
      &= \sum_{n \geq p} K_{n+1} - K_n \\
      &\leq \sum_{n \geq p} 2^{-(n+1)} \\
      &= 2^{-p}
      \end{aligned} 
   \end{equation*}
   for any $p \geq 1$, which concludes the proof of the theorem.
   
   \subsection{Proof of Proposition~\ref{prop:sufficient_stop}} \label{app:proof1}
   
   We first prove that, if there are $w_1, w_2 \in \mathcal{W}$ such that $P_{w_1, w_2} = 1$, then 
   \begin{equation} \label{eq:first_claim_proof_simulation}
     F([w_1]_S) \subseteq [w_2]_S.
   \end{equation}
   Let $w_1$, $w_2$, $k$, $w_1'$ and $w_2'$ be as in Definition~\ref{def:abstraction}. Let us note that 
   \begin{equation} \label{eq:relation_proof}
   \begin{aligned}
     F([w_1]_S) &= \left\{
        F(x) \in X \, \middle| \, 
        \begin{array}{l}
           H(x) = a_1, \\
           H(F(x)) = a_2 \\
           \dots \\
           H(F^{n-1}(x)) = a_{n_1}
        \end{array}
     \right\} \\
     &= F([(a_1)]_S) \cap [(a_2, \dots, a_{n_1})]_S
   \end{aligned}
   \end{equation}
   Now, since $P_{w_1, w_2} > 0$, then $w_1' = w_2'$. There are two cases, either $w_1' = (a_2, \dots, a_{k+1})$ and $w_2' = w_2$, or $w_1' = w_1$ and $w_2' = (b_1, \dots, b_k)$. Let us investigate these separately.
   In the first case, $(a_2, \dots, a_{k+1}) = w_2$. By definition, 
   \begin{equation*}
     [(a_2, \dots, a_{n_1})]_S \subseteq [(a_2, \dots, a_{k+1})]_S = [w_2]_S.
   \end{equation*}
   Therefore \eqref{eq:relation_proof} implies
   \begin{equation*}
     F([w_1]_S) \subseteq F([(a_1)]_S) \cap [w_2]_S \subseteq [w_2]_S, 
   \end{equation*}
   which is \eqref{eq:first_claim_proof_simulation}. In the second case, assume that  
   \begin{equation} \label{eq:to_prove}
     [w_1]_S = [a_1w_2]_S, 
   \end{equation}
   then 
   \begin{equation*}
   \begin{aligned}
     F([w_1]_S)  &= F([a_1w_2]_S) \\
                 &= F([(a_1)]_S) \cap [w_2]_S \\
                 &\subseteq [w_2]_S, 
   \end{aligned}
   \end{equation*}
   where a very similar as in \eqref{eq:relation_proof} was used. It remains to show that \eqref{eq:to_prove} holds. Since we are in the second case cited above, then
   \begin{equation*}
     a_1w_2 = w_1(b_{k+1}, \dots, b_{n_2}), 
   \end{equation*}
   which implies that $[a_1w_2]_S \subseteq [w_1]_S$. Moreover, $P_{w_1, w_2} = 1$, that is 
   \begin{equation*}
     \lambda_X([a_1w_2]_S) = \lambda_X([w_1]_S).
   \end{equation*}
   Following Assumption~\ref{ass:no_manifold}, it means that $[a_1w_2]_S = [w_1]_S$, which proves \eqref{eq:first_claim_proof_simulation}.
   
   Now we prove that \eqref{eq:first_claim_proof_simulation} implies $\mathcal{B}(\Sigma_{\mathcal{W}}) = \mathcal{B}(S)$. We first prove that $\mathcal{B}(S) \subseteq \mathcal{B}(\Sigma_{\mathcal{W}})$. Let $w^* = (a_1, a_2, \dots) \in \mathcal{A}^*$ such that $w^* \notin \mathcal{B}(\Sigma_{\mathcal{W}})$, then there are $a_i, a_{i+1}$ such that, for all $w_1, w_2 \in \mathcal{W}$ for which $L(w_1) = a_i$ and $L(w_2) = a_{i+1}$, $P_{w_1, w_2} = 0$. This could mean three things.
   \begin{enumerate}
     \item For all such $w_1$, $\lambda_X([(a_i)]_S) = 0$. Following Assumption~\ref{ass:no_manifold}, it means that $[(a_i)]_S = \emptyset$.
     \item Let $w_2 = (a_{i+1}, b_2, \dots, b_{n_2})$. For all such $w_2$, 
     \begin{equation*}
        \lambda_X([(a_i, a_{i+1}, b_2, \dots, b_{n_2})]_S) = 0, 
     \end{equation*}
     which means by Assumption~\ref{ass:no_manifold} that 
     \begin{equation*}
        [(a_i, a_{i+1}, b_2, \dots, b_{n_2})]_S = \emptyset.
     \end{equation*}
     By Definition~\ref{def:partitioning}, it implies that $[a_ia_{i+1}]_S = \emptyset$.
   \end{enumerate}
   In any case, it means that $w^* \neq \mathcal{B}(S)$. Now we prove $\mathcal{B}(\Sigma_W) \subseteq \mathcal{B}(S)$. Let $w^* = (a_1, a_2, \dots) \in \mathcal{B}(\Sigma_{\mathcal{W}})$. It means that there exists $w_1, w_2, \dots \in \mathcal{W}$ such that $L(w_i) = a_i$ and $P_{w_i, w_{i+1}} = 1$. Following \eqref{eq:first_claim_proof_simulation}, this implies that $F([w_i]_S) \subseteq [w_{i+1}]_S$. This implies that $[a_ia_{i+1}]_S \neq \emptyset$, which means $w^* \in \mathcal{B}(S)$. 
   
   \subsection{Proof of Corollary~\ref{cor:algo2}} \label{app:proof2}
   Proposition~\ref{prop:sufficient_stop} gives a sufficient condition to stop the algorithm, hence the second part of the claim. It remains to prove that the computational complexity is the claimed one. Let $\mathcal{W}^{(k)}$ and $\mathcal{W}^{'(k)}_i$ be the abstractions $\mathcal{W}$ and $\mathcal{W}_i'$ at iteration $k$ in Algorithm~\ref{alg:adaptive}. First we note that 
   \begin{equation} \label{eq:size_problems_refine}
     |\mathcal{W}^{(k)}| 
     = k|\mathcal{A}| - (k - 1), \quad
     |\mathcal{W}^{'(k)}_i| = (k+1)|\mathcal{A}| - k
   \end{equation}
   At each iteration $k$, one has to compute $|\mathcal{W}^{(k)}|$ times the $\varepsilon$-accurate Kantorovich distance between two models of sizes given by \eqref{eq:size_problems_refine}. By Corollary~\ref{cor:algo}, such computational complexity is
   \begin{equation*}
   \begin{aligned}
     &\mathcal{O}\left(|\mathcal{W}^{(k)}|\left(|\mathcal{A}|^{n+1}\left(|\mathcal{W}^{(k)}|^2 + |\mathcal{W}^{'(k)}_i|^2\right)\right)\right) \\
     = \, & \mathcal{O}\left(|\mathcal{A}|^{n+1} |\mathcal{W}^{'(k)}_i|^3\right) \\
     = \, & \mathcal{O}\left(|\mathcal{A}|^{n+4} (k+1)^3 \right).
   \end{aligned}
   \end{equation*}
   Now, the worst-case is when the algorithm does not converge to an abstraction where there exists $w, w'$ such that $P_{w, w'} \in (0, 1)$. Therefore, the total computational complexity is 
   \begin{equation*}
     \sum_{k = 1}^N \mathcal{O}\left(|\mathcal{A}|^{n+4} (k+1)^3 \right) 
     = \mathcal{O}\left(|\mathcal{A}|^{n+4}N^4\right), 
   \end{equation*}
   which is the claim.
}

\end{document}